\pdfoutput=1

\documentclass[11pt]{article}

\usepackage[]{acl}
\usepackage{times}
\usepackage{times}
\usepackage{latexsym}
\usepackage{colortbl}
\usepackage{inconsolata}
\usepackage{latexsym}
\usepackage{times}
\usepackage{helvet}
\usepackage{courier}
\usepackage{amsfonts}
\usepackage{algorithm}
\usepackage{algorithmic}
\usepackage{amssymb}
\usepackage{multirow}
\usepackage{subcaption}
\usepackage{graphicx}
\usepackage{amssymb}
\usepackage{natbib}
\usepackage{amsmath}
\usepackage{appendix}
\usepackage{hyperref}
\usepackage{booktabs}
\usepackage{graphicx}
\usepackage{makecell}
\usepackage{color}
\usepackage{pifont}
\usepackage{amsfonts}
\usepackage{enumerate}
\usepackage{inconsolata}
\usepackage{diagbox}
\usepackage{times}
\usepackage{latexsym}
\usepackage{tabularx}
\usepackage[T1]{fontenc}

\usepackage[utf8]{inputenc}

\usepackage{microtype}

\title{Knowledge Acquisition Disentanglement for Knowledge-based Visual Question Answering with Large Language Models}

\author{
Wenbin An$^{1}$, Feng Tian$^{1}$, Jiahao Nie$^{2}$, Wenkai Shi$^1$, Haonan Lin$^1$, Yan Chen$^1$, QianYing Wang$^{3}$  \\ \bf{Yaqiang Wu$^{3}$, Guang Dai$^{4}$, Ping Chen$^{5}$} \\
$^1$ Xi'an Jiaotong University $^2$ Nanyang Technological University \\
$^3$ Lenovo Research $^4$ SGIT AI Lab $^5$ University of Massachusetts Boston \\
\texttt{wenbinan@stu.xjtu.edu.cn, fengtian@mail.xjtu.edu.cn} \\ 
\texttt{jiahao007@e.ntu.edu.sg, wangqya@lenovo.com, ping.chen@umb.edu} \\
}

\begin{document}
\maketitle
\begin{abstract}
\textit{Knowledge-based Visual Question Answering (KVQA)} requires both image and world knowledge to answer questions. 
Current methods first retrieve knowledge from the image and external knowledge base with the original complex question, then generate answers with \textit{Large Language Models (LLMs)}. However, since the original question contains complex elements that require knowledge from different sources, acquiring different kinds of knowledge in a coupled manner may confuse models and hinder them from retrieving precise knowledge. Furthermore, the ``forward-only'' answering process fails to explicitly capture the knowledge needs of LLMs, which can further hurt answering quality.
To cope with the above limitations, we propose \textbf{DKA}: \textit{Disentangled Knowledge Acquisition from LLM feedback}, a training-free framework that disentangles knowledge acquisition to avoid confusion and uses LLM's feedback to specify the required knowledge.
Specifically, DKA requires LLMs to specify what knowledge they need to answer the question and decompose the original complex question into two simple sub-questions: \textit{Image-based sub-question} and \textit{Knowledge-based sub-question}. Then we use the two sub-questions to retrieve knowledge from the image and knowledge base, respectively. In this way, two knowledge acquisition models can focus on the content that corresponds to them and avoid disturbance of irrelevant elements in the original complex question, which can help to provide more precise knowledge and better align the knowledge needs of LLMs to yield correct answers. Experiments on benchmark datasets show that DKA significantly outperforms SOTA models. To facilitate future research, our data and code are available at \url{https://github.com/Lackel/DKA}.

\end{abstract}

\begin{figure}
\centering
\includegraphics[width=0.47\textwidth]{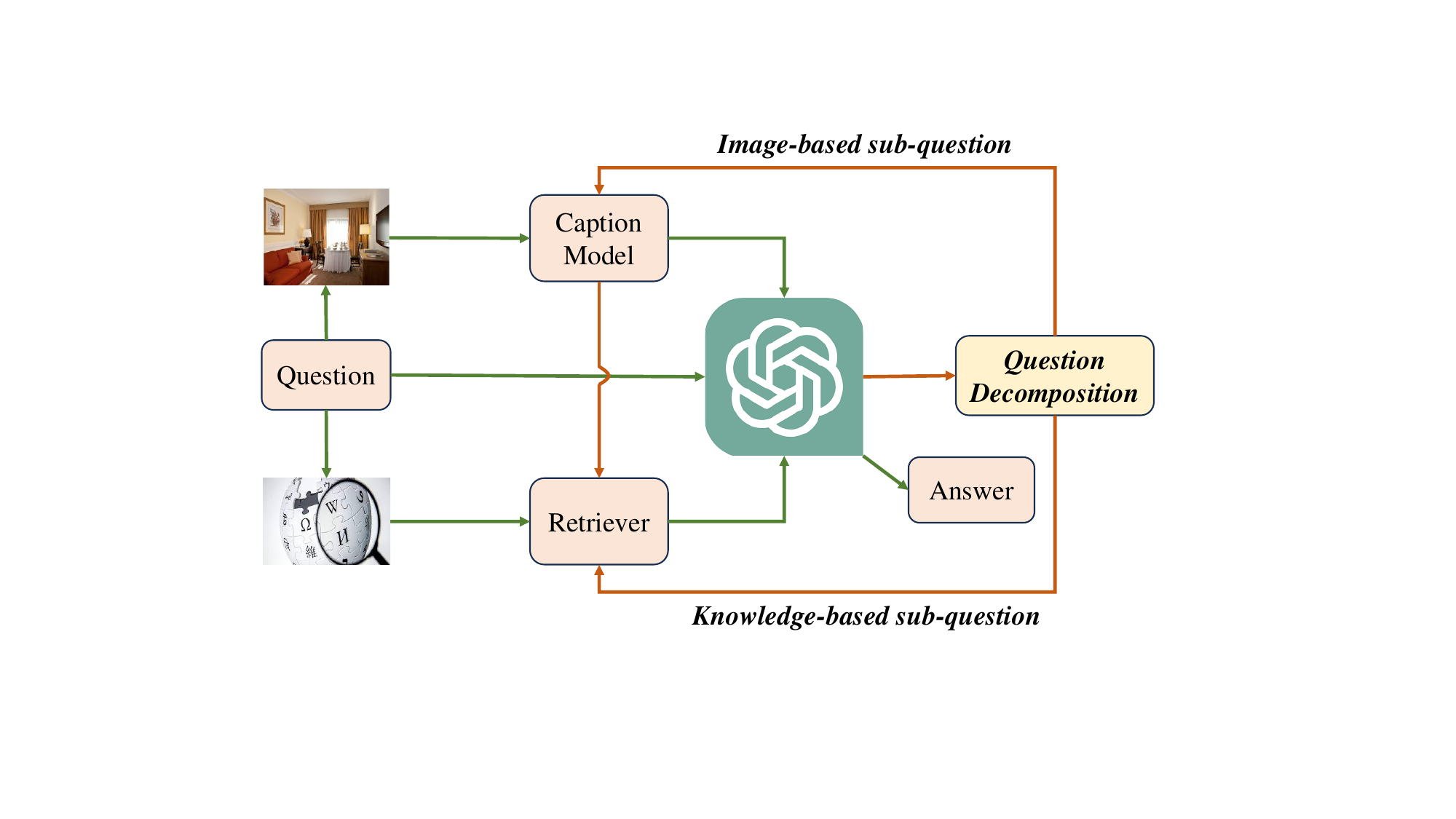}
\caption{\textbf{Green lines:} previous methods acquire knowledge in a coupled manner with only forward process. \textbf{Brown lines:} our model disentangles knowledge acquisition by introducing LLM feedback.} 
\label{fig1}
\end{figure}

\section{Introduction}
\textit{Knowledge-based Visual Question Answering (KVQA)} \citep{okvqa} is a challenging task that requires world knowledge and commonsense reasoning beyond images to answer questions. Benefiting from the rapid development of \textit{Large Language Models (LLMs)}, current state-of-the-art methods \citep{promptcap,heuristics,simple,zero} all use frozen LLMs (e.g., GPT-3 and LLaMA) to answer questions, with image captions and retrieved knowledge as inputs. Specifically, they first retrieve knowledge from the image (e.g., captions \citep{promptcap} and tags \citep{empirical}) and external knowledge base (e.g., Wikipedia \citep{ravqa} and web search \citep{web}) with the original question, then generate answers with LLMs based on the input question and retrieved knowledge (Green lines in Fig. \ref{fig1}).

Despite the improved performance, these LLM-based methods usually struggle to retrieve precise knowledge due to the following two reasons. First, retrieving different kinds of knowledge with the original complex question in a coupled manner can introduce much noise and confuse retrieval models (e.g., caption models and knowledge retriever) since the original question contains complex elements that require knowledge from different sources. For example, using the question \textit{``What \textbf{animal} has similar color to the \textbf{flower} in the image?''} as a query may make the caption model focus on the wrong entity \textbf{animal} in the image and make the knowledge retriever retrieve external knowledge about the wrong entity \textbf{flower}, which inevitably hurts the quality of the retrieved knowledge. 
However, let us consider how humans would solve this problem. Humans would first identify the color of the flower from the image and then retrieve animals with the same color from the knowledge base in a decoupled manner step by step, which will not be interfered by irrelevant information.
Second, these methods contain only a forward process by guessing what knowledge is needed for LLMs but fail to explicitly specify the knowledge needs of LLMs, which can result in a mismatch between retrieved and needed knowledge.

\begin{figure*}
\centering
\includegraphics[width=13.6cm]{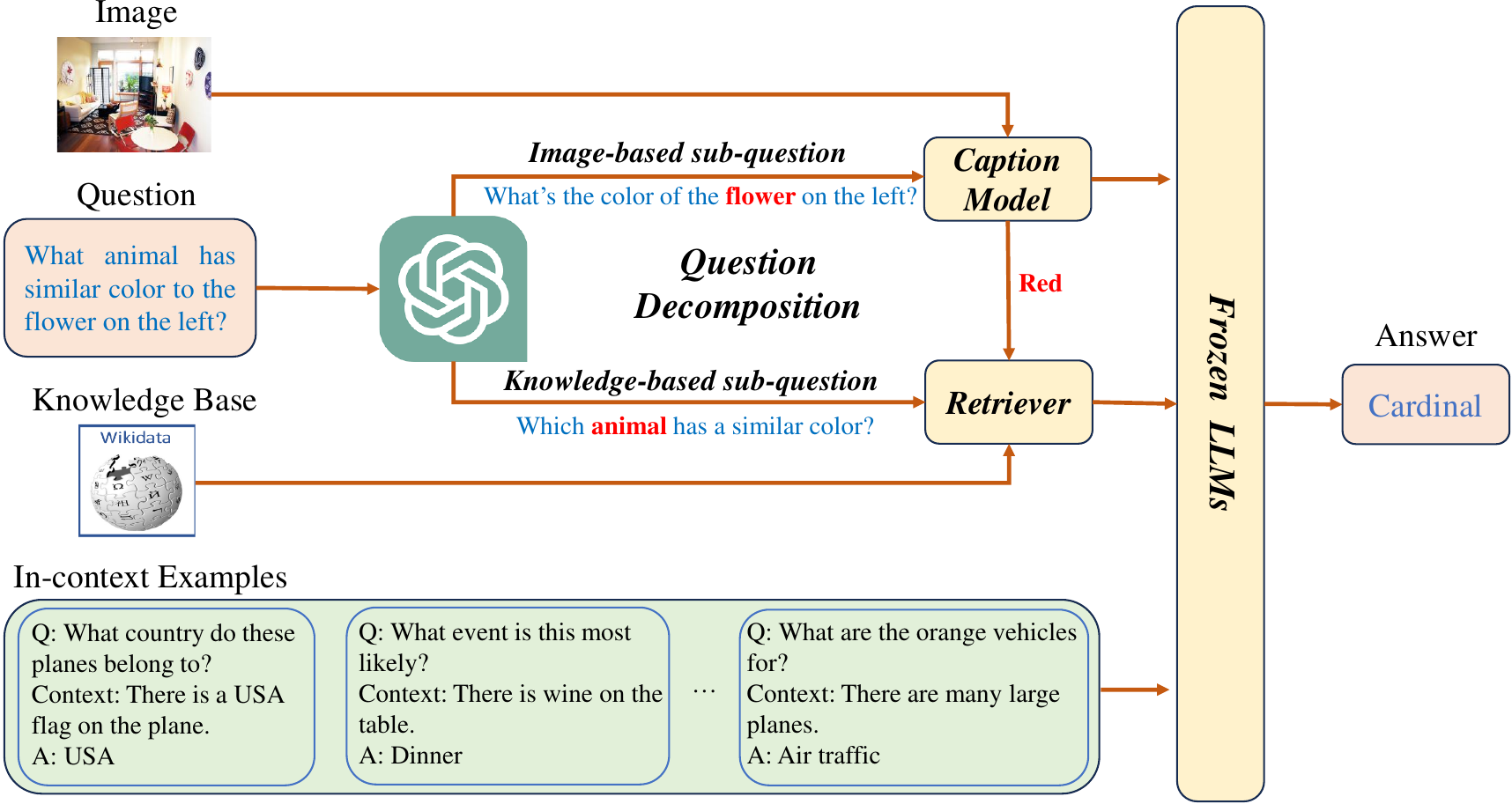}
\caption{The overall architecture of our model.} 
\label{fig2}
\end{figure*}

To cope with the above issues, we propose \textbf{DKA}: \textit{Disentangled Knowledge Acquisition from LLM feedback}, a training-free framework that disentangles knowledge acquisition to avoid confusion and uses LLM's feedback to specify the required knowledge (Brown lines in Fig. \ref{fig1}), which helps to reduce noise and provide more precise knowledge to answer questions. Specifically, inspired by the decoupled thinking process of human beings, DKA requires LLMs to explicitly specify what knowledge they need to answer the question and decompose the original complex question into two simple sub-questions: \textit{Image-based sub-question} and \textit{Knowledge-based sub-question}. Then we use the two sub-questions to retrieve knowledge from the image and knowledge base, respectively. For example, the original question \textit{``What \textbf{animal} has similar color to the \textbf{flower} in the image?''} will be disentangled into \textit{``What’s the color of the \textbf{flower}?''} to query the caption model and \textit{``Which \textbf{animal} has a similar color?''} to query the knowledge base. To incorporate image information, we also input generated captions to the knowledge retriever. In this way, two knowledge acquisition models can focus on the content that corresponds to them and avoid interference of irrelevant entities in the original question, which can help to provide more precise knowledge from different sources and better align the knowledge needs of LLMs to yield correct answers. Experiments on benchmark datasets show that DKA outperforms SOTA methods.

Our main contributions can be summarized as follows:
\begin{itemize}
  \item We propose \textbf{DKA}: \textit{Disentangled Knowledge
   Acquisition from LLM feedback}, a training-free framework for \textit{KVQA}, which disentangles complex questions to avoid confusion and better align the knowledge needs of LLMs.
  \item We propose to disentangle the knowledge acquisition process by decomposing the complex question into two simple sub-questions: \textit{Image-based sub-question} and \textit{Knowledge-based sub-question}, which can help to avoid interference and retrieve precise knowledge.
  \item Extensive experiments show that our model outperforms SOTA methods, achieving 62.1\% and 59.9\% accuracy on the OK-VQA \citep{okvqa} and AOK-VQA \citep{aokvqa} datasets, respectively.
\end{itemize}

\section{Related Work}
\subsection{Training-based KVQA methods}
Training-based KVQA methods usually need training data to train a multimodal encoder to process input images and questions and a language decoder to generate answers \citep{krisp,trg}. For example, KAT \citep{kat} utilizes both explicit and implicit knowledge to generate answers with T5 \citep{t5}. Flamingo \citep{flamingo} trains an 80B Visual Language Models (VLM) for KVQA. CROP \citep{crop}, PNP \citep{pnp} and REVIVE \citep{revive} employ cropped images, GradCAM \citep{gradcam} and regional objects to train a multimodal model to capture the local visual features, respectively. Furthermore, EnFoRe \citep{entity} incorporates local entities for knowledge retrieval to improve KVQA systems. RA \citep{ravqa} uses the output of the decoder to train a knowledge retriever for better knowledge retrieval.

\subsection{Training-free KVQA methods}
Despite the improved performance, training-based KVQA methods rely on abundant labeled data, and the learned knowledge is hard to generalize to different questions. To cope with these limitations, a series of works employ frozen LLMs to answer questions without model training \citep{see,ask}. For example, PICa \citep{empirical} applies GPT-3 with image captions to generate answers. Prophet \citep{heuristics} utilizes vanilla VQA models \citep{mcan} to generate answer heuristics for LLMs. MoQAGPT \citep{moqagpt} uses a multimodal retriever to retrieve multimodal knowledge for LLMs. PromptCAP \citep{promptcap} employs LLMs to generate pseudo labels to train a question-aware caption model. LAMOC \citep{feedback} utilizes LLMs to rank generated captions to refine the pre-trained caption models. Furthermore, Select \citep{select} proposes to generate multiple candidates and then select one answer from them. Decompose \citep{decomposition} proposes to answer many sub-questions to get the final answer \citep{chain}. For example, they decompose the original question \textit{``Does this animal pose a threat to me?''} to sub-questions \textit{``Does the animal have sharp canine teeth?''} and \textit{``Does the animal have forward eyes typical of a predator?''}, relying on manual decomposition and answers for these sub-questions. In contrast, our model can automatically decompose the question and does not require manual labeling for the decoupled sub-questions, which is more efficient.

\section{Method}
\subsection{Approach Overview}
Given an input image-question pair $x=(x_{i}, x_{q})$, the goal of KVQA is to generate the correct answer $y$ for the question $x_{q}$, given the image $x_{i}$ and an external knowledge base $\mathcal{D}$. A training-free LLM-based model can be formalized by maximizing the following probability:
\begin{equation}
    \underbrace{p(c|x_{q}, x_{i})}_{Caption} \cdot \underbrace{p(k|x_{q}, \mathcal{D})}_{Knowledge} \cdot \underbrace{p(y|x_{q}, c, k)}_{Answering}
\end{equation}
where $c$ is the caption generated from question-aware caption models \citep{promptcap,pnp} and $k$ is the retrieved knowledge.
However, as analyzed before, the original question $x_{q}$ contains diverse elements that require knowledge from different sources, which may confuse retrieval models (e.g., caption models and knowledge retriever). To retrieve relevant knowledge more precisely, we propose to disentangle the knowledge acquisition process and reformulate the KVQA task to maximize the following probability:

\begin{small}
\begin{equation}
\label{formulation}
    \underbrace{p(q_{i}, q_{k}|x_{q})}_{Disentangle.} \cdot \underbrace{p(c|q_{i}, x_{i})}_{Caption} \cdot \underbrace{p(k|q_{k}, \mathcal{D}, c)}_{Knowledge} \cdot 
    \underbrace{p(y|x_{q},c,k,e)}_{Answering}
\end{equation}
\end{small}
where $q_{i}$ is an image-based sub-question for caption generation and $q_{k}$ is a knowledge-based sub-question for knowledge retrieval, $e$ is few-shot in-context learning examples. We also add the generated captions for knowledge retrieval to incorporate image information. The overall architecture of our model is shown in Fig. \ref{fig2}. Next, we will introduce the components of our model in detail.

\subsection{Question Decomposition}
To specify the knowledge needs of LLMs and avoid interference of irrelevant entities in the original question, we propose to utilize LLMs to explicitly specify what knowledge they need and decompose the original question into two sub-questions: \textit{Image-based sub-question} for caption models and \textit{Knowledge-based sub-question} for knowledge retriever, which can help to provide more precise knowledge from different sources and better align knowledge needs of LLMs to yield correct answers.
\begin{equation}
    \{q_{i}, q_{k}\} = LLM(\mathcal{P}, x_{q})
\end{equation}
where $\mathcal{P}$ is a prompt template shown in Table \ref{table_prompt}.

\begin{table}[t!]
\centering 
\caption{The prompt for question decomposition.}
\begin{tabularx}{\columnwidth}{|X|}
\hline 
\rowcolor{cyan!20} \texttt{Prompt: Question Decomposition} \\ 
\hline
To answer the question [\textit{Original Question $x_{q}$}] from an image, you should decouple the question into two sub-questions. \\
\\
One sub-question should guide a question-aware caption model to acquire information from the image. \\
\\
Then based on the information from the image, the other sub-question should acquire information from an extra knowledge base. \\
\\
You should return only two questions without explanation in a JSON format. \\
\\
\textbf{Image-based sub-question $q_{i}$:} \\
\\
\textbf{Knowledge-based sub-question $q_{k}$:}
\\
\hline
\end{tabularx}

\label{table_prompt}
\end{table}

\subsection{Question-aware Caption Generation}
Since generic image captions \citep{coco} often miss details essential for LLMs to answer questions, we use the image-based sub-question $q_{i}$ to guide the question-aware caption model PromptCAP \citep{promptcap} to generate informative captions, which can capture visual details about the input question and avoid interference of irrelevant elements:
\begin{equation}
    c = PromptCAP(q_{i}, x_{i})
\end{equation}

\subsection{External Knowledge Retrieval}
Since KVQA requires external world knowledge to answer the question, we further employ a knowledge retriever model to retrieve $r$ relevant knowledge from the external knowledge base $\mathcal{D}$, based on the generated caption $c$ and the disentangled knowledge-based sub-question $q_{k}$:
\begin{equation}
    k = Retriever(q_{k}, \mathcal{D}, c)
\end{equation}

Despite the retrieved knowledge, we further add the local visual information as knowledge by generating 50 image captions with the PNP model \citep{pnp}, which can identify the most related image patches to the question and generate captions only for these patches. While retrieving more relevant knowledge can provide more information, it can also introduce much noise since not all retrieved knowledge is helpful in answering the question. To better balance the quantity-quality trade-off, we further employ the BLIP model \citep{blip} to re-rank cosine similarity between the knowledge and the image $x_{i}$ to capture image-knowledge correlation, and we only keep the top-$n$ most similar knowledge to reduce retrieval noise.

\subsection{In-context Learning}
As previous work \citep{incontext,empirical} suggests, adding some in-context examples to LLMs can yield better performance. Simple \citep{simple} also suggests that selecting in-context examples that are more similar to the query sample can further boost model performance. 
To search for the best in-context examples for each test sample from the training data, we employ the text encoder of MCAN \citep{mcan} to obtain their question similarity and employ the visual encoder of MCAN to obtain their image similarity. Then we use the average of the question and image similarity as a selection criterion following previous work \citep{empirical,simple}. We select the top-$m$ similar samples as in-context samples $e$, which will be fed to LLMs together with the test sample for inference.

\subsection{Answer Generation}
Given original question $x_{q}$, disentangled caption $c$, knowledge $k$, and selected in-context samples $e$, we feed them together into LLMs to generate the answer prediction $y$. Specifically, at each decoding step $t$, the generated token $y^{t}$ can be decided by the following probability:
\begin{equation}
    y^{t} = \mathop{argmax}_{y^{t}}p_{LLM}(y^{t}|x_{q},c,k,e,y^{<t})  
\end{equation}
where $LLM$ represents the weights of large language models, which are frozen during inference.

\begin{table*}[ht]
\centering
\caption{Statistics of benchmark datasets. \# Q, Avg. Q length, Avg. A length represent the number of questions, average question length, and average answer length, respectively. Statistics are based on the results reported by AOK-VQA \citep{aokvqa}.}
\begin{tabular}{lccccc}
\hline
Dataset & \# Q & Knowledge Type & Avg. Q length  & Avg. A length & Unique Words \\
\hline
OK-VQA      &  5,046     &  Factoid       & 8.1   & 1.3  &5,703\\
AOK-VQA     &  1,145 + 6,702     &  Common/World  & 8.8   & 1.3  &7,248\\
\hline
\end{tabular}
\label{table111}
\end{table*}

\begin{table*}[ht]
\setlength\tabcolsep{10pt}
\centering
\caption{
Comparison with SOTA methods on the OK-VQA dataset. Large language models are frozen during inference. Some results are cited from previous works \citep{simple,heuristics}.
}
\begin{tabular}{lccc}
\toprule[1.2pt]
Method & Image Representation & Model & Accuracy (\%)\\
\midrule
\multicolumn{4}{c}{\textbf{End-to-End Training}}\\
Mutan & Feature & ResNet + GRU & 26.4 \\
Mucko & Feature & LSTM & 29.2 \\
ConceptBERT & Feature & BERT & 33.7 \\
KRISP &  Feature & MMBERT & 38.4 \\
Vis-DPR   &  Feature  & LXMERT  & 39.2 \\
MAVEx  & Feature & ViLBERT & 39.4  \\
TRiG  & Caption + Tags + OCR & T5-large & 50.5 \\
KAT (Single)  & Caption + Tags + Feature & T5-large & 53.1   \\
KAT (Ensemble)  & Caption + Tags + Feature & T5-large & 54.4   \\
REVIVE (Single)  & Caption + Feature & T5-large & 56.6 \\
REVIVE (Ensemble)  & Caption + Feature & T5-large & 58.0 \\
\midrule
\multicolumn{4}{c}{\textbf{In-Context Learning \& Zero-Shot Learning}} \\
PNP-VQA (zero-shot) & Caption &  UnifiedQAv2 (11B) & 35.9 \\
LAMOC (zero-shot)  & Caption &  FLAN-T5-XXL (11.4B) & 40.3 \\
PICa-Base (In-context)  & Caption + Tags &  GPT-3 (175B) & 43.3   \\
BLIP-2 (zero-shot) & Feature & FlanT5-XXL (11B) & 45.9 \\
PICa-Full (In-context) & Caption + Tags &  GPT-3 (175B) & 48.0   \\
Flamingo (80B) & Feature &  Chinchilla (70B) & 57.8 \\ 
PromptCap (In-context) & Caption  & GPT-3 (175B) & 60.4 \\
Prophet (In-context) & Caption  & GPT-3 (175B) & 61.1 \\
Simple (In-context) & Caption  & LLaMA 2 (13B) & 61.2 \\
\rowcolor{gray!20} \textbf{DKA (Ours)} & Caption  & LLaMA 2 (13B) & \textbf{62.1} \\
\bottomrule[1.2pt]
\end{tabular}

\label{tab:okvqa}
\end{table*}

\subsection{Answer ensemble}
Prompting LLMs $q$ times to obtain $q$ answer predictions and ensemble them to get the final answer can help to yield more robust and better performance \citep{empirical,simple}. So we also ensemble $q$ answer predictions and select the one with the highest sum of output log-probability as the final answer:
\begin{equation}
    y = \mathop{argmax}_{i=\{1,2,...,q\}}\sum_{t}logp_{LLM}(y_{i}^{t})
\end{equation}
where $p_{LLM}(y_{i}^{t})$ represents the output probability of the $i$-th answer at the step $t$.

\section{Experiments}
\subsection{Dataset}
\textbf{OK-VQA} \citep{okvqa} contains 5,046 image-question pairs for testing, where questions are manually filtered to ensure that outside knowledge is required to answer the question.

\noindent \textbf{AOK-VQA} \citep{aokvqa} contains 1,145 image-question pairs for validation and 6,702 image-question pairs for testing, where each question is annotated with ten answers for evaluation.

\noindent More details about the benchmark dataset OK-VQA and AOK-VQA are listed in Table \ref{table111}. From the table we can see that compared to the OK-VQA dataset, the AOK-VQA dataset has more questions, requires more diverse and complex knowledge, and has a more diverse vocabulary of questions, which makes the dataset more challenging.

\subsection{SOTA Methods for Comparison}
We compare our model with both end-to-end training-based methods and in-context learning or zero-shot learning training-free methods.

\noindent \textbf{Training-based methods.} Mutan \citep{mutan}, ConceptBERT \citep{conceptbert}, KRISP \citep{krisp}, Vis-DPR \citep{dpr}, ClipCap \citep{aokvqa}, Pythia \citep{pythia}, MAVEx \citep{mavex}, TRiG \citep{trg}, KAT \citep{kat} and REVIVE \citep{revive}.

\noindent \textbf{Training-free methods.} PNP-VQA \citep{pnp}, LAMOC \citep{feedback}, PICa \citep{empirical}, BLIP-2 \citep{blip2}, ViLBERT \citep{vilbert}, LXMERT \citep{lxmert}, GPV-2 \citep{gpv2}, Unified-IO \citep{unified}, Flamingo-80B \citep{flamingo}, PromptCap \citep{promptcap}, Prophet \citep{heuristics} and Simple \citep{simple}.

\subsection{Implementation Details}
Following previous work, we use the official VQA accuracy \citep{okvqa} as the evaluation metric. For knowledge acquisition disentanglement, we query frozen LLaMA2-13B \citep{llama2} with the prompt in Table \ref{table_prompt}. Then we employ PromptCAP \citep{promptcap} (470M parameters) to generate captions given an image and an image-based sub-question. For external knowledge retrieval, we use ChatGPT to retrieve $r$ relevant knowledge items. Furthermore, we also employ PNP-VQA \citep{pnp} to generate 50 local captions as knowledge, following previous work \citep{simple}. For in-context learning, we utilize MCAN \citep{mcan} to select the most informative examples following previous work \citep{promptcap,simple}. Finally, we use frozen LLaMA2-13B \citep{llama2} to generate answers. For hyper-parameters, we follow the setting in Simple \citep{simple} to make a fair comparison. The number of selected knowledge items $n$ is set to 9, the number of in-context examples $m$ is set to 10, the number of answer ensemble $q$ is set to 5, and the number of retrieved knowledge items $r$ is set to 10. All experiments are conducted on an NVIDIA A40 GPU.

\begin{table}[t]
\centering
\caption{Comparison with SOTA methods on the AOK-VQA dataset.}
\setlength\tabcolsep{8pt}
\begin{tabular}{lcc}
\toprule[1.2pt]
Model & Val Acc (\%) & Test Acc (\%) \\
\midrule
ClipCap      &  18.1     &  15.8                \\
Pythia       &  25.2     &  21.9                \\
ViLBERT      &  30.6     &  25.9                 \\
LXMERT       &  30.7     &  25.9                 \\
CRISP        &  33.7     &  27.1                 \\
GPV-2        &  48.6     &  40.7                 \\
Unified-IO   &  -        &  45.2                \\
Prophet      &  58.2     &  55.7                \\
PromptCap    &  56.3     &  59.6                \\
Simple       &  58.6     &  57.5                \\
\rowcolor{gray!20}\textbf{DKA (Ours)}       &  \textbf{62.1}     &  \textbf{59.9} \\
\bottomrule[1.2pt]
\end{tabular}
\label{tab:aokvqa}
\end{table}

\subsection{Main Results}
\noindent \textbf{Comparison results on OK-VQA.} Table \ref{tab:okvqa} summarizes the comparisons of our model DKA with SOTA methods on the OK-VQA dataset. Our model outperforms both training-based and LLM-based methods and achieves state-of-the-art performance. Specifically, our model outperforms the SOTA training-based method REVIVE \citep{revive} by 4.1\% and outperforms the SOTA LLM-based method Simple \citep{simple} by 0.9\%, which shows the effectiveness of our model. Furthermore, compared to GPT3-based models, our model achieves better performance with LLaMA 2 \citep{llama2}, an open-source model with much fewer parameters than GPT-3, which can be deployed locally to protect data privacy and save on the cost of API queries. Compared to large vision language models (e.g., Flamingo \citep{flamingo}), our model does not require massive multimodal data and computing power to train the model, which can reduce training costs.

\noindent \textbf{Comparison results on AOK-VQA.} Table \ref{tab:aokvqa} summarizes the comparisons of our model DKA with SOTA methods on the validation and testing sets of the AOK-VQA dataset. Our model outperforms all compared methods and achieves state-of-the-art performance on both validation and testing sets. Specifically, our model outperforms the SOTA method Simple \citep{simple} by 3.5\% on the validation set and outperforms the SOTA method PromptCap \citep{promptcap} by 0.3\% on the testing set, which shows the effectiveness of our model.

\noindent \textbf{Summary.} Our model DKA achieves the state-of-the-art performance on three datasets. We contribute the improvement to two reasons. First, DKA can align knowledge retrieval with the knowledge needs of LLMs by utilizing the feedback of LLMs, which can help retrieve more useful knowledge. Second, DKA disentangles the knowledge acquisition process to avoid interference of irrelevant entities in the original question, which can make models focus on the content that corresponds to them and retrieve more precise knowledge from different sources.

\section{Discussion}
\subsection{Ablation Study}
The performance of variants of our model on the AOK-VQA dataset is shown in Table \ref{tab:ablation}.

\noindent \textbf{Disentangled Knowledge Acquisition.} 
(1) Removing disentangled external knowledge (\textbf{w/o Disentangled Knowledge}) can hurt model performance since answering some questions requires knowledge beyond the ability of LLMs. However, this effect is relatively small because LLMs are competent to answer most questions that require common sense knowledge. However, if the question requires up-to-date knowledge beyond the ability of LLMs, we still need to retrieve knowledge from the external knowledge base.
(2) Removing disentangled captions (\textbf{w/o Disentangled Caption}) can largely hurt model performance since captions are responsible for providing global image information for LLMs.
(3) Removing both disentangled knowledge and captions (\textbf{w/o Knowledge + Caption}) has the greatest impact on model performance since KVQA requires both image information and external knowledge to answer questions.

\noindent \textbf{In-context Example Selection.} We also investigate the effects of different in-context example selection strategies on model performance. (1) Using MCAN \citep{mcan} (\textbf{w/ MCAN}) as the example selector yields the best performance since the model has the ability to capture similarities between different image-question data pairs. (2) Using BLIP \citep{blip2} (\textbf{w/ BLIP}) as the example selector also yields good performance because of the multimodal capability of BLIP. (3) However, selecting in-context examples randomly (\textbf{w/ Random}) can largely hurt model performance, which means that selecting similar in-context examples is important for KVQA \citep{simple}.

\begin{table}[t]
\centering
\caption{Ablation study on the AOK-VQA dataset.}
\setlength\tabcolsep{4pt}
\begin{tabular}{lc}
\toprule[1.2pt]
Model & Val Acc (\%)  \\
\midrule
\rowcolor{gray!20}\textbf{DKA (Ours)}       &  \textbf{62.1}     \\
\midrule
\multicolumn{2}{c}{\textit{\textbf{Disentangled Knowledge Acquisition}}} \\
w/o Disentangled Knowledge  &  61.3 \\
w/ Original Question        &  60.7 \\
w/o Disentangled Caption    &  58.1 \\
w/o Knowledge + Caption     &  57.5 \\
\midrule
\multicolumn{2}{c}{\textit{\textbf{In-context Example Selection}}} \\
w/ MCAN   &  62.1 \\
w/ BLIP   &  60.7 \\
w/ Random &  59.3 \\
\bottomrule[1.2pt]
\end{tabular}
\label{tab:ablation}
\end{table}

\begin{table*}
\caption{Hyper-parameter experiments on the AOK-VQA validation dataset.}
\setlength\tabcolsep{15pt}
  \begin{subtable}{0.23\textwidth}
    \begin{tabular}{lc}
    \toprule
    $q$ & Acc (\%)  \\
    \midrule
    1   &  59.9 \\
    3   &  61.9 \\
    5   &  62.1 \\
    7   &  61.2 \\
    9   &  61.6 \\
    \bottomrule
    \end{tabular}
    \subcaption{The number of answer ensemble $q$.}
    \label{ensemble}
  \end{subtable}
  \hspace{0.23cm}
  \begin{subtable}{0.23\textwidth}
    \begin{tabular}{cc}
    \toprule
    $m$ & Acc (\%)  \\
    \midrule
    1   &  57.3 \\
    3   &  60.8 \\
    5   &  61.5 \\
    10  &  62.1 \\
    15  &  61.3 \\
    \bottomrule
    \end{tabular}
    \subcaption{The number of in-context examples $m$.}
    \label{shot}
  \end{subtable}
  \hspace{0.23cm}
  \begin{subtable}{0.23\textwidth}
    \begin{tabular}{cc}
    \toprule
    $r$ & Acc (\%)  \\
    \midrule
    1   &  61.6 \\
    3   &  62.0 \\
    5   &  61.7 \\
    7   &  61.7 \\
    10  &  62.1 \\
    \bottomrule
    \end{tabular}
    \subcaption{The number of retrieved knowledge $r$.}
    \label{retrievek}
  \end{subtable}
  \hspace{0.23cm}
  \begin{subtable}{0.23\textwidth}
    \begin{tabular}{cc}
    \toprule
    $n$ & Acc (\%)  \\
    \midrule
    1   &  60.1 \\
    3   &  61.3 \\
    9   &  62.1 \\
    12  &  61.5 \\
    15  &  61.3 \\
    \bottomrule
    \end{tabular}
    \subcaption{The number of selected knowledge $n$.}
    \label{selectk}
  \end{subtable}
  \label{tab:hyper}
\end{table*}

\begin{table}[t]
\centering
\caption{Comparison with Large Vision Language models (LVLMs) on the metric VQA accuracy (\%). It should be noted that most of these models are pre-trained on the OKVQA and AOKVQA datasets.}
\begin{tabular}{lcc}
\toprule[1.2pt]
Model & OKVQA & AOKVQA \\
\midrule
MiniGPT-4      &  52.5     &  54.2                \\
QWen-VL       &  58.6     &  -                \\
Shikra      &  47.2     &  -                 \\
InstructBLIP (T5)       &  55.5     &  54.8                 \\
InstructBLIP (Vicuna)        &  62.1     &  62.1                 \\
LLaVA 1.5        &  63.3     &  -                 \\
DKA (Ours)       &  62.1     &  59.9                \\
\bottomrule[1.2pt]
\end{tabular}
\label{tab:lvlms}
\end{table}

\subsection{Hyper-parameter Analysis}
\noindent \textbf{Effect of answer ensemble.} Table \ref{ensemble} shows the accuracy with different number of answer ensembles $q$. From the table, we can see that employing multiple prompts to get multiple answers and then ensemble them yields improved accuracy over a single answer because the ensemble strategy allows the model to utilize more in-context examples for inference and reduce the impact of output randomness. However, the retrieval of irrelevant examples can also cause the fluctuation of accuracy when $q$ is large \citep{simple}.

\noindent \textbf{Effect of in-context learning.} Table \ref{shot} shows the effects of different number of in-context examples $m$. From the table, we can see that our model gets comparable performance even with 1 shot example. And as the number of in-context examples grows, our model gets better performance. However, when $m$ is large, the retrieval of irrelevant in-context examples can also cause the fluctuation of accuracy, which is similar to the case of the answer ensemble.

\noindent \textbf{Effect of retrieved knowledge.} Table \ref{retrievek} shows the effects of different numbers of retrieved knowledge $r$ from the external knowledge base. From the table, we can see that our model is not sensitive to the change of $r$, which is because the most relevant knowledge can be maintained since they have higher similarity to the question, and our knowledge selection strategy can keep this relevant knowledge and filter out other irrelevant disturbances.

\noindent \textbf{Effect of retrieved knowledge.} Table \ref{selectk} shows the effects of different numbers of selected knowledge $n$, which contains both the retrieved knowledge and the local image knowledge generated by PNP \citep{pnp}.
From the table, we can see that when $n$ is small, the accuracy increases with $n$ since more knowledge relevant to the question can provide better generalizability for LLMs. However, when $n$ is large, the accuracy begins to fluctuate since the selected knowledge may contain much noise that has low similarity to the question and affects the output of LLMs.

\begin{figure}[t]
\centering
\includegraphics[width=7.5cm]{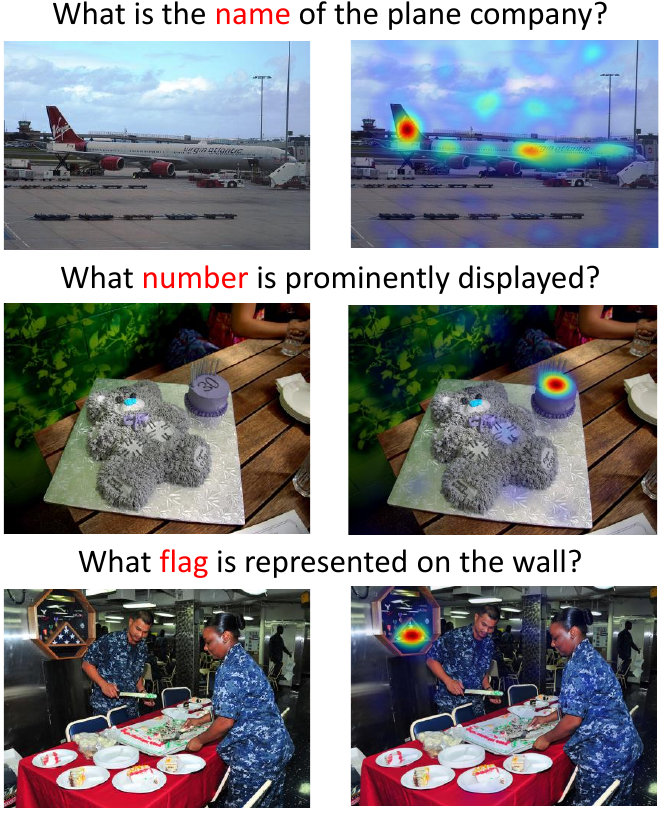}
\caption{Heatmap visualization with GradCAM.} 
\label{fig5}
\end{figure}

\begin{figure*}[t]
\centering
\includegraphics[width=16.1cm]{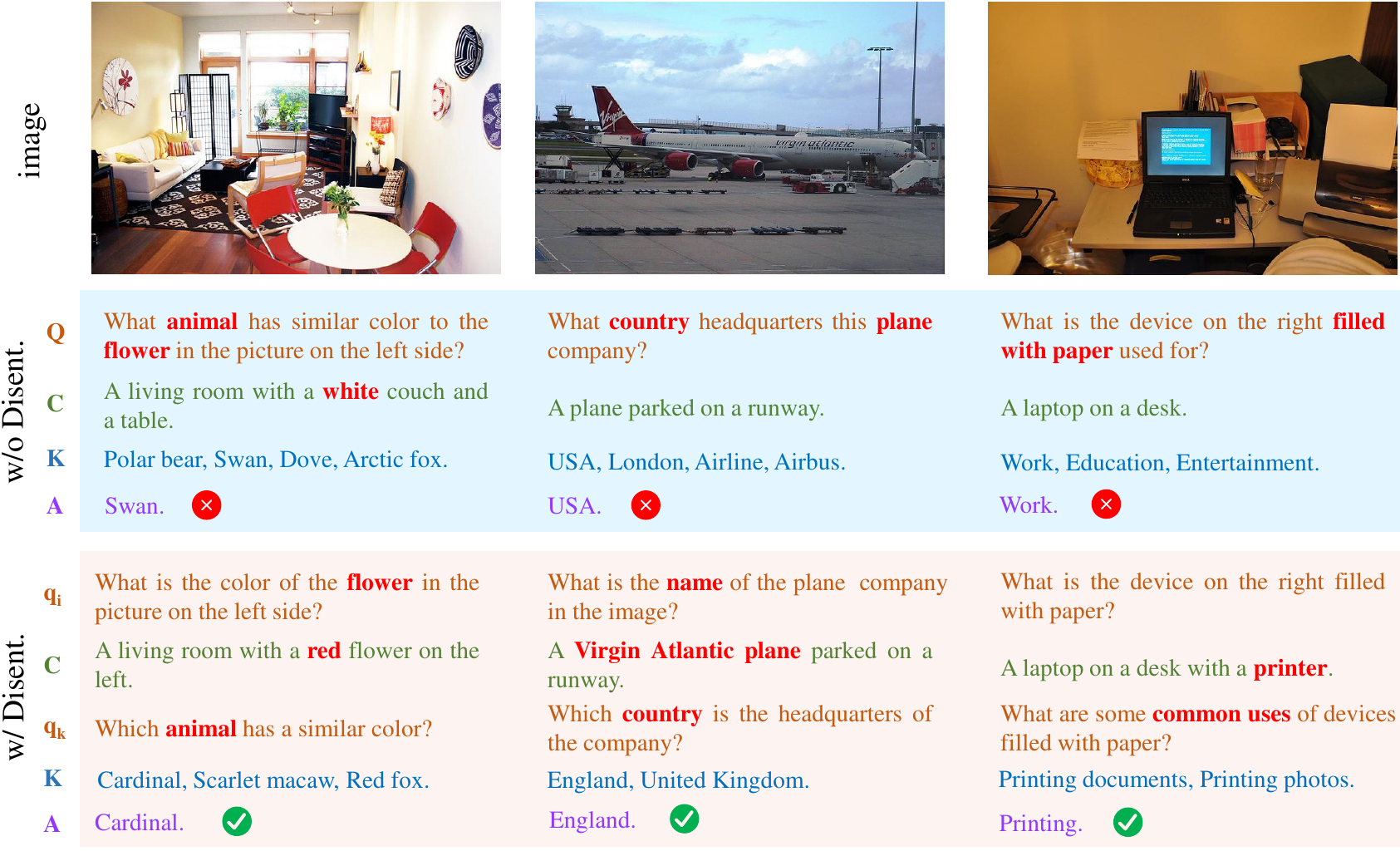}
\caption{Examples with (w/) and without (w/o) disentanglement. $Q$, $C$, $K$, $A$, $q_{i}$, $q_{k}$ represent the question, caption, retrieved knowledge, answer, image-based sub-question, knowledge-based sub-question, respectively. Some important elements are marked in red.} 
\label{fig3}
\end{figure*}

\subsection{Comparison with Large Vision Language models}
Large Vision-Language Models (LVLMs) have demonstrated strong ability in visual recognition and language understanding, so we also compare our model with recent state-of-the-art LVLMs. Specifically, we compare our model with MiniGPT-4 \citep{minigpt}, Qwen-VL \citep{qwen}, Shikra \citep{shikra}, InstructBLIP \citep{instructblip} and LLaVA \citep{llava}. As shown in Table \ref{tab:lvlms}, our model with only a few demonstration examples gets comparable performance to these models that are pre-trained on the entire set of OKVQA \citep{okvqa} and AOKVQA \citep{aokvqa}, which can show the effectiveness of our model. Furthermore, compared to LVLMs, LLM-based models have the following additional advantages for Knowledge-based Visual Question Answering (KVQA). First, LLM-based models can directly utilize well-trained caption models and LLMs without requiring additional data and computing resources to train and align these models, which is more efficient. Second, compared to the coupled inference way (without generating explicit captions) of LVLMs, LLM-based methods reason in a decoupled manner, which has higher interpretability. If an error occurs, we can easily locate the wrong position and change a better caption model, a better knowledge retriever, or a larger LLM, without needing to re-train the entire model to align different modules. However, for LVLMs, if we want to replace the visual encoder or the LLM, we need to re-train them to align between images and text. Last, without the captions, it will be hard for LVLMs to retrieve knowledge from the knowledge base due to the lack of visual information, which may limit their applications on the knowledge-centric tasks (e.g., KVQA). In summary, LLM-based models and LVLMs may adapt to different needs and application scenarios: LVLMs are more suitable for vision-centric tasks like traditional VQA because of their strong ability in visual understanding, and LLM-based models may be more suitable for knowledge-centric tasks like complex knowledge-based VQA since they are more suitable for knowledge retrieval and have better interpretability.

\subsection{Heatmap Visualization}
Fig. \ref{fig5} shows some examples of the GradCAM heatmap \citep{gradcam,pnp} guided by disentangled image-based sub-questions. From the figure we can observe that disentangled sub-questions can help to capture the most relevant regions in the image and ignore other irrelevant interference so that the caption model \citep{pnp,promptcap} can focus on these question-relevant regions without interference and generate question-related captions to help LLMs to yield correct answers for the given complex questions.
\vspace{-1mm}

\subsection{Qualitative Analysis}
Fig. \ref{fig3} illustrates some examples generated with and without disentanglement. From the figure, we can see that disentangled questions help to avoid the disturbance of irrelevant entities in the original question so that the caption model can focus on the most relevant regions in the image and generate question-related captions. Furthermore, the generated captions can help to provide precise image information for knowledge retrieval. In summary, our \textit{Disentangling Knowledge Acquisition (DKA)} model can utilize the feedback of LLMs to disentangle an original question into two sub-questions and make two knowledge acquisition models focus on the corresponding content, which can avoid interference and provide more precise knowledge for LLMs to generate correct answers.

\begin{figure*}[t]
\centering
\includegraphics[width=0.98\textwidth]{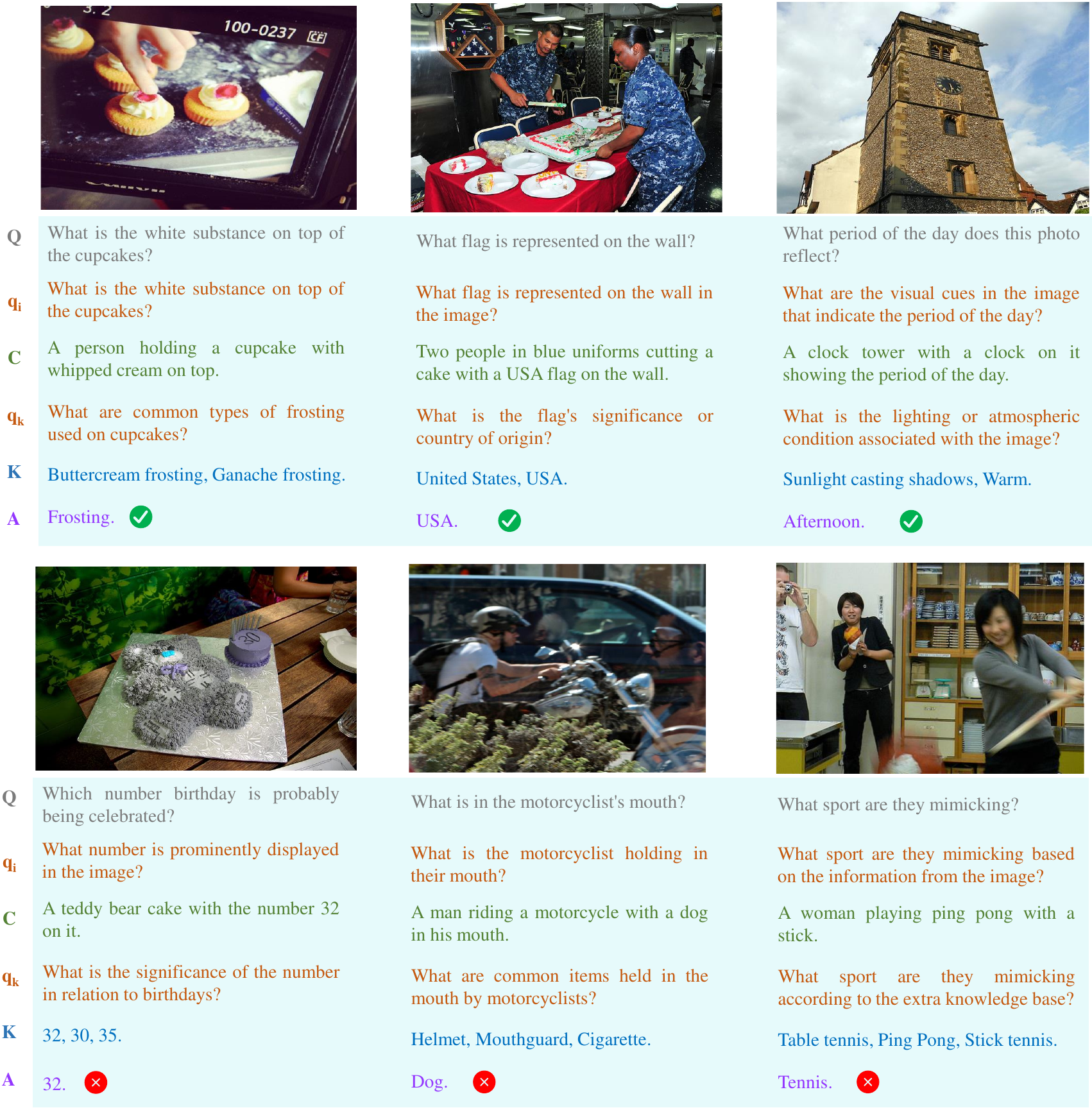}
\caption{Examples with disentanglement. $Q$, $C$, $K$, $A$, $q_{i}$, $q_{k}$ represent the question, caption, retrieved knowledge, answer, image-based sub-question, knowledge-based sub-question, respectively. }
\label{fig6}
\vspace{-2.5mm}
\end{figure*}

\subsection{Examples and Error Analysis}
Fig. \ref{fig6} illustrates more examples of the disentangled knowledge acquisition. From the figure, we can see that by decoupling the original complex questions into easy sub-questions, DKA can help avoid the disturbance of irrelevant entities and generate more accurate answers.
Furthermore, we also provide some error examples in the last line of Fig. \ref{fig6}. From these examples, we can see that the error mainly comes from the insufficient ability of the caption model, which can not provide correct visual information about the image because of multimodal hallucinations. So how to adjust the caption model with the help of LLMs to yield more correct captions will be our future research direction.

\section{Conclusion}
In this paper, we propose \textit{Disentangling Knowledge Acquisition (DKA)} for \textit{KVQA}, a training-free few-shot framework that disentangles the knowledge acquisition process to avoid confusion and utilizes LLM’s feedback to specify the required knowledge. 
By decomposing an original complex question into two simple sub-questions: \textit{Image-based sub-question} and \textit{Knowledge-based sub-question} with LLM feedback, \textit{DKA} can make the two knowledge acquisition models focus on the content that corresponds to them and avoid interference of irrelevant elements in the original complex question, which can help to capture more accurate visual information, retrieve more precise external knowledge and better align the knowledge needs of LLMs. Experiments on two benchmark datasets show that our model with a few demonstration examples outperforms previous SOTA LLM-based methods and gets comparable performance to SOTA LVLMs that are pre-trained on the entire set of OKVQA and AOKVQA datasets, which can validate the effectiveness of our model. 

\vspace{-1mm}

\section*{Limitations}
Even though the proposed model achieves superior performance on the KVQA task, it still faces the following limitations.
First, it is important to acknowledge that we have not yet explored the performance of using other open-source LLMs for question decomposition and answering because of the high computation and time overhead. 
Second, even though our model enjoys the efficiency of zero training and achieves improved performance, we acknowledge that training component models (e.g., caption models, knowledge retrievers, and answering models) with LLM feedback may result in better performance. So how to better balance training efficiency and model performance will be our future research direction.

\section*{Acknowledgments}
This work was supported by National Science and Technology Major Project (2022ZD0117102), National Natural Science Foundation of China (62293551, 62177038, 62277042, 62137002, 61937001,62377038). Project of China Knowledge Centre for Engineering Science and Technology, ‘‘LENOVO-XJTU’’ Intelligent Industry Joint Laboratory Project.

\bibliography{anthology}
\end{document}